# Design for a Darwinian Brain: Part 1. Philosophy and Neuroscience

March 2013

*Chrisantha Fernando*

Dept. of Electronic Engineering and Computer Science
Queen Mary University of London

**Abstract**

Physical symbol systems are needed for open-ended cognition. A good way to understand physical symbol systems is by comparison of thought to chemistry. Both have systematicity, productivity and compositionality. The state of the art in cognitive architectures for open-ended cognition is critically assessed. I conclude that a cognitive architecture that evolves symbol structures in the brain is a promising candidate to explain open-ended cognition. Part 2 of the paper presents such a cognitive architecture.

**Introduction**

This is the first of a two-part paper that aims to fuel our imagination about how a physical symbol system with the properties defined by Fodor and Pylyshyn could plausibly exist and be refined by learning in the brain. I do this by making an analogy between biochemical and neuronal processes. The genetic regulatory system of the cell is a physical symbol system that is grounded in the world through cell signaling networks and metabolism. It is the result of stochastic optimization undertaken by natural selection at phylogenetic timescales. It is a chemical symbol system to be exact. Analogously, I propose two plausible candidates for a neuronal symbol system. The first consists of spatiotemporal patterns of spikes that are operated on by re-write rules implemented as neuronal transformations. Another candidate consists of patterns of connectivity between neuronal groups. The first proposal has been explored previously (Fernando 2011) and the second proposal is described at the cognitive level in Part 2 of this paper. The theory of Darwinian neurodynamics proposes biologically plausible neuronal mechanisms by which these symbol structures can replicate in the brain, and thus undertake a process of stochastic optimization of the neuronal symbol system during development, for example to implement language acquisition or action grammar. Once a plausible basis for symbol systems has been proposed, it is necessary to understand how that system functions to cause behavior, i.e. how it is grounded. This has also been the approach of molecular biology that seeks to understand the meaning of a gene, and how that meaning arises, i.e. to find out what a gene "is for".

The paper begins with a description of cells as chemical symbol systems, and then describes one way in which neuronal symbol systems could be implemented in the brain. It concludes by considering the algorithmic functional possibilities for how a neuronal symbol system could be coupled to the non-symbolic apparatus of the brain to produce behavior. This final question is of-course the most challenging, just as in systems biology, even having a complete genome has not told us how a cell works, we should expect the same of recent approaches to the connectome.

Tibor Ganti describes the concept of fluid automata, or chemical machines (Gánti 2003). Chemical machines are made of molecules. Molecules are objects composed of atoms that have specific structural relations to other atoms in the molecule. A molecule is assembled according to what in terms of physical symbol system talk would be referred to as a combinatorial syntax, i.e. a set of chemical structural constraints such as valance, charge, etc… that determine how atoms can 'legally' join together to make a molecule. I would go further and say that combinatorial semantics is then what determines how a molecule with a particular structure will react (behave) in a given environment. So, semantic content in terms of a chemical symbol system equates to its chemical function, or in other words its reactivity. The function of a molecule is itself a function of the semantic content of its parts, translated; the reactivity of a benzene ring is modified by its side-groups such as methyl groups. In short, the chemical symbols and their structural properties *cause* the system behavior.

This chemical symbol system operates in parallel rather than in series. It is constrained by kinetics. Its function (reactivity) is subjected to non-encoded influences such as temperature that influences the thermodynamic equilibrium position of chemical reactions. These aspects do not tend to intrude in normal conversations about physical symbol systems, but they are germane to chemical symbol systems. A pleasing example of a symbolically specified computation in chemistry is a chemical clock. The two autocatalytic cycles of the Belousov-Zhabotinsky (BZ) reaction constitute a fluid automaton that implements a chemical clock. Whilst it is the symbolic organization of the molecules that specifies the reaction network topology, it is at the level of the analog (dynamical systems) operations of the assembled reaction network that the clock like phenomenon of the BZ reaction exists. Continuous behavior results from the ensemble properties of a multitude of discrete symbolic chemical operations. The tremendous success of chemistry is to have robustly linked the chemical system level with the macroscopic level.

Of-course, it is the fact that molecules are chemical symbols that give chemistry this rich set of macroscopic characteristics in the first place. For example, chemistry has **productivity**. Productivity means that a system can encode indefinitely many propositions, that there is recursive assembly of parts of the representation to produce the proposition, and that an unbounded number of non-atomic representations is entailed (Fodor and Pylyshyn 1988). If we replace the term proposition with adaptation, then in biological evolution, the number of possible adaptations is unlimited in the same sense, being the developmental result of a symbolic genetic and epigenetic encoding. Whether it is helpful to replace proposition with adaptation is a moot point for philosophers who argue about representations, but to me this seems natural because adaptations have a form of truth-value in the sense that they are the entities to which fitness applies. Similarly, the capacity for chemical reactivity is unlimited, i.e. there are many more possible chemical reactions than could be implemented in any realistically sized system. An unlimited number of molecules can be produced allowing an unlimited many chemical reactions, and this is made possible with only a finite set of distinct atom types. Therefore, an unbounded set of chemical structures must be composite molecules. In physical symbol systems terms it is also the case that an indefinite number of propositions can be entertained, or sentences spoken. This is known as the productivity of thought and language. If neural symbol systems exist, they must have the capacity of being combined in unlimited ways. The next section deals with how this could potentially work. Finally, there is reason to believe that sadly no non-human animal has the capacity for productive thought (Penn, Holyoak et al. 2008) and the differences between other primates and us may tell us how neuronal symbol systems function, a subject we return to at the end of this paper.

The second interesting property of chemistry is that it is **systematic,** meaning that the capacity for atoms to be combined in certain ways to produce some molecules is intrinsically connected to their ability to produce others. This is neatly shown by how chemists learn chemistry. There are several heuristics or rules of thumb that chemists can learn to help them predict properties of molecules and how they will react based on their structure. Only a chemist quite lacking in insight will attempt to rote learn a list of valid reactions. It is in a similar way that there is systematicity in language. The ability to produce and understand a sentence is intrinsically connected with the ability to produce and understand other sentences. Languages are not learned by learning a phrasebook, they have syntax which allows the systematic combination of entities.

The third property of chemistry is that an atom of the same type makes roughly the same contribution to each molecule in which it occurs. For example, the contribution of a hydrogen atom to a water molecule influences many reactivity properties of that molecule. Hydrogen atoms donate electrons and this is a property of the hydrogen atom itself, whatever in most cases it binds to. This means there is systematicity in reactivity (semantics) as well as in structure (syntax). This is known as **compositionality.** In the same way, lexical items in sentences have approximately the same contribution to each expression in which they occur. This approximate nature suggests that there is a more fundamental set of 'atoms' in language than words themselves, e.g. linguistic constructions.

Why was the idea of a chemical symbol system even entertained in chemistry? Why did people come to believe in discrete atoms coming together systematically to form molecules? Lavoisier discovered a systematic relationship between chemical reactions, i.e. the conservation of mass. Proust discovered the law of definite proportions, i.e. that compounds can be broken down to produce constituents in fixed proportions. Dalton extended this law of multiple proportions that explained that when two elements come together to form different compounds (notably the oxides of metals), they would come together in different small integer proportions. Are there any analogous reasons to believe in neuronal symbol systems, based on observation of human thought and language?

Certainly more is demanded of a human neuronal physical symbol system. Open-ended cognition appears to require the ability to learn a new symbol system, not just to implement an already evolved one. Children can learn and manipulate explicit rules (Marcus 2001) and this implies the existence of a neuronal symbol system capable of forming structured representations and learning rules for operating on those representations. For example, Gary Marcus has shown that 7 month old infants can distinguish between sound patterns of the form ABA versus ABB, where A and B can consist of different sounds e.g. "foo", "baa" etc. Crucially, these children can generalize this discrimination capacity to new sounds that they have never heard before, as long as they are of the form ABA or ABB. Marcus claims that performance in this task requires that the child must extract "abstract algebra-like rules that represent relationships between placeholders (variables), such as "the first item X is the same as the third item Y", or more generally that "item I is the same as item J". Several attempts have been made to explain the performance of these children without a PSS (e.g. using connectionist models) but Marcus has criticized these as smuggling in symbolic rules in one way or another by design. For Marcus it seems that the system itself must discover the general rule. In summary, the problem with a large set of connectionist learning devices is that a regularity learned in one component of the solution representation is not applied/generalized effectively to another part. Marcus calls this the problem of training independence. Marcus considers this one of the fundamental requirements for a learning system to be described as symbolic or rule based.

The following is a concise definition of a symbol system adapted from Harnad to emphasize the chemical aspects (Harnad 1990). A symbol system contains a set of arbitrary atoms (or physical tokens) that are manipulated on the basis of "explicit rules" that are likewise physical tokens or strings (or more complex structures, e.g. graphs or trees) consisting of such physical tokens. The explicit rules of chemistry generate reactions from the structure of atoms and molecules (plus some implicit effects, e.g. temperature). The rule-governed symbol-token manipulation is based purely on the shape of the symbol tokens (not their "meaning"), i.e., it is purely syntactic, and consists of "rulefully combining" and recombining symbol tokens, in chemical reactions. There are primitive atomic symbol tokens and composite symbol-token strings (molecules). The entire system and all its parts – the atomic tokens, the composite tokens, the syntactic manipulations both actual and possible and the rules – are all "semantically interpretable :" The syntax can be systematically assigned a meaning e.g., as standing for objects or as describing states of affairs. For example, semantic interpretation in chemistry means that the chemical system exhibits chemical reactivity. In biochemistry this is extended to a higher level. In the chemicals inside cells, it is fair to say that properties of those chemicals, e.g. the concentration, or their configuration, actually stands for states of affairs in the environment outside the cell, for example the conformation of a signaling molecule may represent the glucose concentration outside the cell. In the same way a neural symbol system exhibits behavior such as the child's capacity to distinguish ABA from ABB in grammar learning tasks.

This analogy has helped me to understand what Fodor and Pylyshyn were talking about in their famous paper denouncing connectionism (Fodor and Pylyshyn 1988). From this basis I developed what I believe may be two plausible candidates for a neuronal symbol system. The first has been described in detail in two previous publications (Fernando 2011; Fernando 2011). In short it consists of arbitrary physical tokens (spatiotemporal patterns of spikes) arranged into molecules or symbol structures (groups of spatiotemporal patterns of spikes passing through the same network). These tokens undergo explicit rule-governed symbol-token manipulation (reactions involving transformation of the spike patterns), and a process of stochastic optimization can learn the explicit rule sets. Thus the proposed neural representation of an atomic symbol token is a spatiotemporal patterns of spikes. Neurons detect a particular spatiotemporal spike pattern if the axonal delays from the pre-synaptic neuron to the detector neuron are properly matched to the spatiotemporal pattern such that depolarization reaches the detector neuron body simultaneously. This implementation of neural symbol-tokens (atoms) uses the concept of polychronous computing and a modification of the concept of wavefront computing (Izhikevich 2006; Izhikevich and Hoppensteadt 2009). The construction of molecular symbol structures from atomic symbol-tokens requires binding of atomic symbol-tokens together such that they can be subsequently manipulated (reacted) as a function of the structure of the molecule. In my framework, compositional neural symbolic structures exist as temporally ordered sequences of symbols along chains of neurons. A symbol-structure can be described as a sequential combination of spike patterns, a feature that has been described in real neuronal networks as a 'cortical song' (Ikegaya and al 2004). I hypothesize that a great many such short chains of combined spatiotemporal spike patterns exist in the brain. Each chain can be considered to be a kind of register in a computer, blackboard or tape that can store symbol-tokens of the appropriate size. A single symbol-token could be read by a detector neuron with the appropriate axonal delay pattern when interfacing with the chain. A parallel symbol system in the brain consisting of a population of such chains, each capable of storing a set of symbol token molecules and operating on these molecules in parallel is conceivable. Interaction between (and within) such chains constitutes the operations of symbol-manipulation. Returning to the chemical metaphor, such interactions can be thought of as chemical reactions between molecules contained on separate chains, and rearrangements within a molecule expressed on the same chain. Whilst in a sense a chain can be thought of as a tape in a Turing machine (due to the serial nature of the strings), it also can be thought of as a single molecule in a chemical system (due to the existence of multiple parallel chains). This constitutes the core representational substrate on which symbol manipulation will act.

Receiving input from the chain and writing activity back into the chain is done by a detector neuron with specific input and output delays in relation to the chain. A detector neuron only fires when the correct pattern of input is detected. In effect the neuron is a classifier. Where the classifier fails to fire, the same pattern enters the chain as leaves the chain. This is because the spatiotemporal organization of these patterns does not match the spatiotemporal tuning curve of the detector neuron. Only when the spatiotemporal spike pattern matches does the detector neuron fire. Once fired, the output of the detector neuron is injected back to the neurons of the chain. If the output of the detector neuron slightly precedes the normal passage of the untransformed pattern through the chain, then the refractory period of the output neurons of the chain prevents interference by the original untransformed pattern, which is thereby replaced by the new pattern specified by the detector neuron. Such a detector neuron we will now call a classifier neuron because it is a simple context free re-write rule with a condition (detection) and an action pole of the type seen in Learning Classifier Systems (LCS) (Holland and Reitman 1977; Wilson 1995).

It can be seen that such classifier neurons are selective filters, i.e. the classifier neuron is only activated if the spatiotemporal pattern is sufficiently matched with the axonel delays afferent upon the neuron. The above classifier implements an implicit rule. An implicit rule is a rule that operates on atomic or molecular symbol structures without being specified (encoded/determined/controlled) by a symbol structure itself. There is no way that a change in the symbol system, i.e. the set of symbols in the population of chains, could modify this implicit matching rule. The implicit rule is specified external to the symbol system. Whenever the symbol passes along this chain, it will be replaced by the new symbol, irrespective of the presence of other symbols in the system.

In a symbol system (as in chemistry), symbols are manipulated (partly) on the basis of "explicit rules ". This means that the operations or reactivity of symbols depends on/is controlled by/is causally influenced by their syntactic and semantic relationship to other symbols within the symbol-structure and between symbol structures. If inhibitory gating neurons are combined with classifier neurons then it becomes possible to implement explicit rules within our framework. If an inhibitory gating unit must receive a particular pattern of spikes in order for it to become active, and if that unit disinhibits a classifier neuron that is sensitive to another symbol token then it is possible to implement a context-sensitive re-write rule. The rule is called context sensitive because the conversion of X to Y depends on the relation of X to another contextual symbol T. A set of context-sensitive re-write rules is capable of generating a grammar of spike-patterns. Consider starting the system off with a single symbol-token S. Probabalistic application of the rules to the initial symbol S would result in the systematic production of spike patterns consisting of grammatically correct context-sensitive spike pattern based sentences. A major implementation issue in real neuronal tissue would be the fidelity of transmission of spatiotemporal spike patterns. The information capacity of such a channel may fall off with decreasing fidelity of copying in that channel in a manner analogous to Eigen's error catastrophe in genetic evolution (Eigen 1971). Finally, we have shown how it is possible to undertake stochastic optimization in the above system by copying with mutation of neuronal classifiers (Fernando 2011). This mechanism depends on spike-time dependent plasiticity (STDP) by which one neuron infers the re-write rule implemented by another neuron, with errors. This permits a population of re-write rules to be evolved. The details of this mechanism are beyond the scope of this paper but can be found elsewhere (Fernando 2011).

We now move onto a much harder topic indeed. What kinds of behavior require a physical symbol system, if any? We've understood something about how chemical and genetic systems generate reactions and cellular behavior. But what about how neural representations produce organismal behavior? Whilst a gene is a functional 'unit of evolution', we can ask, what are the possible functional 'units of thought' at the algorithmic level, i.e. what are the entities in the brain that encode behavioral traits whose probability of transmission during learning is correlated with reward (or value functions of reward) (Price 1970; Price 1995). It may well be the case that there exist units of thought that do not have compositionality, systematicity and productivity at all. For example, a solitary feed-forward neural network that implements a forward model which predicts the next state of the finger as a function of the previous joint angles and the motor command given to the finger, does **not** satisfy the definition of a symbol system discussed above. Such models of the world may be complex, e.g. inverse models of the motor commends required to reach a desired goal state, but may lack the properties of a physical symbol system. The units may contribute to many domains from the linguistic to the sensorimotor, or occupy a level of specialized niches, e.g. units in the cerebellum, or units in the primary visual cortex. By analogy with units of evolution (Maynard Smith 1986) they may be called 'units of thought', although their individuality may be more a convenience for measurement and description than a strict reality, for example it is possible that they are physically overlapping. I define them as the entities in the brain whose survival depends upon value, whatever form such value takes, e.g. extrinsic or intrinsic values (Stout, Konidaris et al. 2005). How far can an animal get in cognition without physical symbol systems? In short the mystery is this; what are the kinds of representation upon which search must act in order to produce cognition, and how do certain classes of representation map to certain classes of cognition? A concrete proposal is given in Part 2.

The second question is what kinds of learning algorithm can operate to improve neural representations? This is no trivial question. And again, this is a question that must be answered whether or not the representations satisfy the definition of a physical symbol system. In answering this question I have found it important to notice that both thought and evolution are intelligent, and undertake knowledge-based search as defined by Newell (Newell 1990). I propose that both open-ended thought and open-ended evolution share similar mechanisms and that a close examination of these two processes in context is helpful. Traditionally thought has been considered to be directed and rational (Newell and Simon 1972), whereas evolution by natural selection has been considered random and blind (Dawkins 1976). Neither view in their extreme forms is correct. A necessary feature of creative thought is stochastic search (Perkins 1995; Simonton 1995) and evolution by natural selection whilst it may be a blind watchmaker is not a stupid one, it can learn to direct its guesses about what variants are likely to be fit in the next generation (Pigliucci 2008). Apparently a grandmaster, maybe Reti, was once asked "How many moves do you look ahead in chess?" to which he answered "One, the right one"! We now know that evolution makes moves (mutants) in the same way, not by explicitly looking ahead, but by recognizing what kinds of moves tended to be good in the past in certain situations and biasing mutation accordingly (Kashtan and Alon 2005; Izquierdo and Fernando 2008; Parter, Kashtan et al. 2008). In fact, this principle has been exploited in the most sophisticated black-box optimization algorithms such as covariance matrix adaptation evolution strategies (CMA-ES) that rivals the most sophisticated reinforcement learning algorithms (Stulp and Sigaud 2012). How do the mechanisms of cognition differ from the mechanisms of genetics in discovering how to make new adaptive ideas/organisms effectively? It seems that processes of systematic inference are in place in both adaptive systems, e.g. bacterial genomes are organized into Operons that through linkage disequilibrium tend to bias variability between generations along certain dimensions, i.e. crossover is a structure sensitive operation, a notion that was coined to describe the Classical cognitive paradigm (Fodor and Pylyshyn 1988). Structure sensitive operations are the symbolic equivalent of methods to bias variation in continuous systems such as CMA-ES.

Both evolution and thought are generative processes with systematicity and compositionality because both depend on structure sensitive operations on representations (Fodor and Pylyshyn 1988) that far exceed the complexity of CMA-ES. We are able to generate 'educated guesses' about hidden causes, to make up explanations for which there is as yet insufficient data to use deduction. This is evidenced by children's ability to learn language (Goldberg 1995; Steels 2006; Steels and De Beule 2006). In fact there is considerable evidence much of which I find convincing, that no non-human animal can entertain models of the world that require the hypothesis of unlimited kinds of hidden causes (Penn, Holyoak et al. 2008). Genetic and epigenetic generative mechanisms seem rather inflexible in comparison to those of thought. From close examination of these abilities can we have some idea what 'units of thought' must be capable of representing about the world, i.e. how they limit the kinds of concept we can entertain? Perhaps we have a chance by looking at an example of constrained creativity from Gestalt psychology in the form of insight problems (Sternberg and Davidson 1995). The 9-dot problem is a task in which you must cover each dot in a 3 x 3 matrix of dots with just 4 straight lines without removing your pen from the paper. Through the application of introspection we seem to solve this problem by generating a set of possible solutions, inventing some kind of fitness function for the solutions such as how many dots they cover, and mutating the better solutions preferentially in a systematic way. We may not be able to verbalize the scoring function nor how the initial set of solutions was chosen or varied. Often we will reach an impasse, a kind of sub-optimal equilibrium in thinking about this problem. The impasse may be diffused by suddenly discovering the correct solution that arises in a punctuated event called the 'Ah-ha' moment. How are the solutions to the above problem represented, how many distinct solutions are represented at once in the brain, how are they varied over time, how do solutions interact with each other, how does a population of solutions determine behavior, and how are the scoring functions represented in the brain, and how many there are? We have proposed elsewhere that the solution of such problems involves a Darwinian search in the brain (Fernando and Szathmáry 2009). It may be the case that careful modeling of the above kinds of thinking process can uncover the kinds of representation and the search algorithm used to search in the space of such representations.

In cognitive science it is useful to put your money where your mouth is and actually try to build a robot to do something. Part 2 shows how we propose to build a robot that has open-ended creativity and curiosity that exhibits productivity, compositionality, and systematicity of thought. The most advanced attempts to build robots capable of open-ended creativity are briefly first. The general principle in all that work has been to learn predictive models of self and world. These are typically implemented as supervised learning devices such as neural networks. The networks are not physical symbol systems. However, some approaches attempt to combine predictors compositionally. The approach that I have decided to take in the remainder of Part 1 is to examine the limitations of these approaches in producing productive behavior in robots.

The approach taken by most groups in trying to achieve open-ended cognition is to learn predictive models of the world that become as rich as possible, and to guide action selection on the basis of how well certain models are being learned. The idea is that as we grow up we try to model the world in greater and greater richness and depth. An early and beautiful approach to this problem was by A.M. Uttley who invented a 'conditional probability computer' and connected it to a Meccano robot that was controlled with a hard-wired Braitenberg type controller (Andrew 1958; Uttley 1959)[1]. This computer learned to associate sensory states with motor actions when this external controller was controlling the robot, and could learn the sensorimotor rules, e.g. turn left if left sensor dark and right sensor light, and eventually substitute this hard-wired controller. Note that this was an entirely associationist form of model and would fall under what Fodor and Pylyshyn call

---

[1] I am ashamed to say that I also rediscovered Uttley's ideas about spatiotemporal spike pattern learning in a recent paper without having known about his work Fernando, C. (2011). "Symbol Manipulation and Rule Learning in Spiking Neuronal Networks." Journal of Theoretical Biology **275**: 29-41.

Connectionism. More recent approaches learn state(t)—motor(t)—next state(t+1) triplets through experience, generating a network of anticipations (Butz, Shirinov et al. 2010). Once such a cognitive map is learned it can be used for model-based reinforcement learning (dynamic programming) to backpropogate rewards associated with goal sensory states to earlier states and thus to permit planning of high-value trajectories. The capacity to simulate sense-action trajectories is a critical function of a cognitive map. The sophistication of mental simulation is a significant constraint on intelligence and creativity. Note that there is no physical symbol system involved in the above algorithm for cognitive map building. How then might a physical symbol system help in learning models of the world?

An obvious limitation of the above kind of cognitive map formation is that it may fail when there are unobserved causes. One solution is the generation of classifiers that infer the existence of hidden-causes (Nessler, Pfeiffer et al. 2010) and then utilize this in further modeling of the world. By this means an unlimited conceptual world opens. This is where physical symbol systems may well enter the picture. Once a space of higher-level classifiers exists, the combinatorial explosion of possible conditions and predictions requires an effective method for sparse search in this higher-order space. The space of higher-order predictions is also highly structured. In summary, the units of thought in TGNG are s(t)-m(t)-s(t+1) triplets that store transition probabilities. There is no need for compositionality and systemeticity of such units because they can be entirely instructed by supervised learning methods (at least in low dimensional sensorimotor systems). Efficient sparse search algorithms would however be needed when hypothesizing hidden states for which there is insufficient sensory evidence to undertake supervised learning. So, by thinking about the problem from this robotic perspective, I have identified a domain in which stochastic processes and representations with compositionality and systematicity would be expected to bring efficiency, i.e. *during search in a structured space of hidden causes*.

Interestingly, a modification of the simple cognitive maps above that potentially adds compositionality and systematicity has been proposed in order to scale reinforcement learning to "domains involving a large space of possible world states or a large set of possible actions" (Botvinick, Niv et al. 2009). The solution has been to chunk actions into higher-level actions consisting of sequences of primitive actions, called 'options' (Sutton, Precup et al. 1999). Options can be hierarchically organized policies. Each abstract action specifies a policy (a map from states to actions) that once started from in initiation state, executes a sequence of actions (which may also be options) until a termination state is reached. They work by reducing the size of the effective search space by allowing structuring of search by modular recombination (concatenation) of options. **The critical question is how the correct set of useful options can be chosen in the first place.** Typically options are hand-designed, with options specifying sub-goals to be achieved. In none of the models are options stochastically explored. The options framework is an admirable first step towards achieving a truly systematic and compositional physical symbol system for describing actions. It is related to the notion of action grammar developed by Patricia Greenfield (Greenfield 1991) which also seeks to identify the behavioural correlates of neural symbol systems.

To conclude, for any reasonable sized sensory and motor dimensionality, and for reasonable time-periods, exhaustive search in predictive model space is not possible.

The computation is exponential in time and space. To solve this problem, units of thought (e.g. predictors and actors) should exhibit compositionality and systematicity, of the kind for example that we see in the options framework. The system should have the ability to form hierarchical descriptions of action and hierarchical predictive models. Part 2 presents our 'fluid option framework' or 'fluid action grammar' in which option-like symbol structures are co-evolved in the brain along with their fitness functions.

**Acknowledgements**

Thanks to a Templeton Grant FQEB "Bayes, Darwin and Hebb" and the FP-7 FET OPEN project INSIGHT-II for funding this work. Thanks to Eors Szathmary and Adam Sanborn, and two anonymous reviewers for comments.

Living Machines 2013 Natural History Museum, London.